\documentclass[final]{cvpr}
\usepackage{times}
\usepackage{epsfig}
\usepackage{graphicx}
\usepackage{amsmath}
\usepackage{amssymb}
\usepackage{amsmath}
\usepackage{booktabs}
\usepackage{multirow}
\usepackage{footnote}
\usepackage{pifont}
\newcommand{\cmark}{\ding{51}}%
\newcommand{\xmark}{\ding{55}}%
\newcommand\norm[1]{\left\lVert#1\right\rVert}

\usepackage{caption}
\usepackage{lipsum}
\usepackage{float}
\usepackage{enumitem}

\usepackage[pagebackref=true,breaklinks=true,colorlinks,bookmarks=false]{hyperref}

\usepackage[accsupp]{axessibility}

\begin{document}

\title{HyperTransformer: A Textural and Spectral Feature Fusion Transformer for Pansharpening\vspace{-1mm}}

\author{Wele Gedara Chaminda Bandara, Vishal M. Patel\\
Johns Hopkins University\\
Department of Electrical and Computer Engineering, Baltimore, MD 21218, USA\\
{\tt\small \{wbandar1, vpatel36\}@jhu.edu}
}

\maketitle

\begin{abstract}
   Pansharpening aims to fuse a registered high-resolution panchromatic image (PAN) with a low-resolution hyperspectral image (LR-HSI) to generate an enhanced HSI with high spectral and spatial resolution. Existing pansharpening approaches neglect using an attention mechanism to transfer HR texture features from PAN to LR-HSI features, resulting in spatial and spectral distortions. In this paper, we present a novel attention mechanism for pansharpening called HyperTransformer, in which features of LR-HSI and PAN are formulated as queries and keys in a transformer, respectively.  HyperTransformer consists of three main modules, namely two separate feature extractors for PAN and HSI, a multi-head feature soft-attention module, and a spatial-spectral feature fusion module. Such a network improves both spatial and spectral quality measures of the pansharpened HSI by learning cross-feature space dependencies and long-range details of PAN and LR-HSI. Furthermore, HyperTransformer can be utilized across multiple spatial scales at the backbone for obtaining improved performance. Extensive experiments conducted on three widely used datasets demonstrate that HyperTransformer achieves significant improvement over the state-of-the-art methods on both spatial and spectral quality measures. Implementation code and pre-trained weights can be accessed at \url{https://github.com/wgcban/HyperTransformer}.
\end{abstract}

\section{Introduction}
    \vspace{-2mm}
    \label{sec:introduction}
    \begin{figure}[tb!]
    	\centering
    	\includegraphics[width=\linewidth]{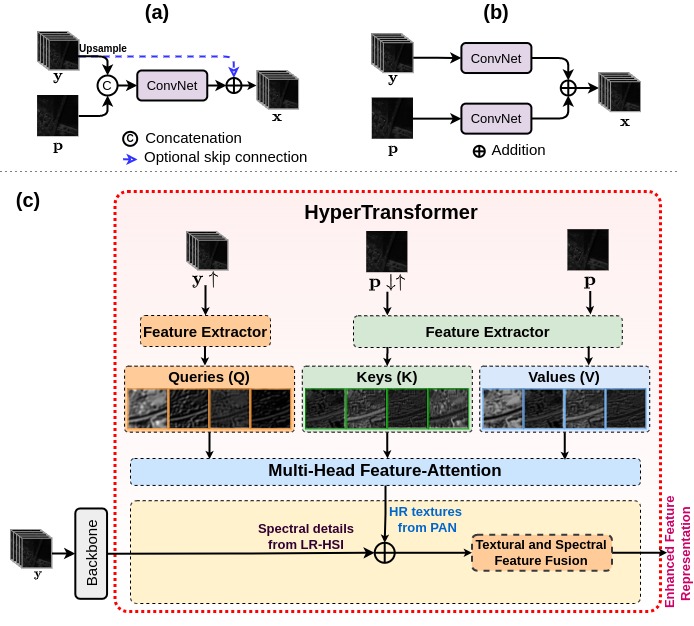}
    \vskip-10pt	\caption{How our \textit{HyperTransformer} differs from existing pansharpening architectures. Traditional pansharpening methods simply concatenate PAN ($\mathbf{p}$) and LR-HSI ($\mathbf{y}$) in \textbf{(a)} image domain~\cite{DHP-DARN, cvpr_2021_sipsa} or \textbf{(b)} feature domain~\cite{Hyper-PNN, wang2021dual, iccv_21_feature_fusion} to learn the mapping function from LR-HSI to pansharpened HSI ($\mathbf{x}$). In contrast,  \textbf{(c)} our HyperTransformer utilizes feature representations of LR-HSI, PAN$\downarrow \uparrow$, and PAN as Queries (Q), Keys (K), and Values (V) in an attention mechanism to transfer most relevant HR textural features to spectral features of LR-HSI from a backbone network.  The output of HyperTransformer is an enhanced version of the feature representation of $\mathbf{y}$. $\uparrow$ and $\downarrow $ denote bicubic upsampling and down-sampling, respectively. \vspace{-6mm}}
    \label{fig:overview_ST}
    \end{figure}
    \par Hyperspectral (HS) pansharpening aims to spatially enhance Low-Resolution Hyperspectral Images (LR-HSIs) by transferring textural (spatial) details from better spatial resolution panchromatic (PAN) images, while preserving the spectral characteristics of LR-HSIs~\cite{HSI_review_new, iccv_2017}. The recent advancements in HS pansharpening greatly improve the amount of spectral and textural details in HSIs, which is indeed a crucial pre-processing  for many remote sensing applications to accurately and rapidly identify the underlying phenomena that would otherwise be difficult to see from LR-HSIs. HS pansharpening can be beneficial in a broad range of remote sensing tasks such as unmixing~\cite{HS_unmixing}, change detection ~\cite{HS_change_d, changeformer}, object recognition~\cite{HS_road_d}, scene interpretation~\cite{HS_scene_i}, and classification~\cite{HS_class,spin}.

    \par The early research on HS pansharpening employed component substitution (CS)~\cite{CS_IHS, CS_PCA, CS_GS}, multi-resolution analysis (MRA)~\cite{MRA_Wavelet, MRA_Lapalacian}, Bayesian~\cite{Bayesian1, Bayesian3}, and variational \cite{Bayesian2, 2019CVPR_variational,yao2020cross} methods to transform spatial details from PAN image to LR-HSI. However, these traditional pansharpening approaches often result in spatial and spectral distortions due to improper modeling of prior knowledge, inaccessibility of sensor characteristics, the mismatch between prior assumptions with the problem~\cite{cvpr_2020_unsup_adaptation} (such as linear spectral mixture assumption~\cite{pan_mixer_assumption} and the sparsity assumption~\cite{pan_sparsity1}), and reliance on hand-crafted features such as  dictionary~\cite{pan_sparsity1, pan_sparsity2} with limited representation ability.

    \par Recently, deep convolutional neural networks (ConvNets) have also been introduced for HS pansharpening due to their excellent ability to learn proper image features. However, state-of-the-art (SOTA) approaches often adopt straightforward ways to transfer textural and spectral details from PAN image to LR-HSI. For example, Lee et al.~\cite{CVPR_21_SPISA}, Zheng et al.~\cite{DHP-DARN} and Bandara et al.~\cite{DIP-HyperKite} adopted a network shown in Figure \ref{fig:overview_ST}-(a) as the backbone to learn the mapping function from the concatenation of up-sampled LR-HSI and PAN to the pansharpened HSI. However, we argue that the concatenation of PAN image along with hundreds of LR spectral bands makes textural and spectral feature fusion difficult, and inefficient. In addition, it could result in high spectral and spatial distortions in pansharpened HSI due to the inappropriate mixing of textural-spectral details. In contrast to the image-domain concatenation, researchers have also investigated feature-domain concatenation of PAN and LR-HSI as shown in Figure \ref{fig:overview_ST}-(b). In this approach, two separate ConvNets are utilized to extract HR textural patterns from PAN, and spectral properties from LR-HSI ~\cite{wang2021dual, Hyper-PNN}. However, still the mixing process of textural and spectral details is just the addition without any appropriate guidance/attention over features. We argue that the above approaches do not effectively utilize the cross-feature space dependency between LR-HSI and PAN, and the long-range details of PAN during the textural-spectral mixing process. Instead, they completely rely upon the succeeding convolutional operations to propagate relevant textural-spectral features through the network. Although the convolution operation with sufficient depth is able to fuse the textural-spectral features appropriately to some extent, it is not intended to adjust each pixel value based on global (long-range) spectral-spatial details of the feature maps, but to adjust values of the small spatial regions together by employing the convolution kernel, which is not accurate and appropriate specially in HS pansharpening.

    \par Motivated by a recent work on image super-resolution~\cite{yang2020learning}, we propose a novel textural-spectral feature fusion transformer called \textit{HyperTransformer} for HS pansharpening that addresses the aforementioned issues of conventional pansharpening approaches as depicted in Figure \ref{fig:overview_ST}-(c). In contrast to conventional pansharpening approaches, our HyperTransformer utilizes an attention mechanism to extract cross-feature space dependency between PAN and LR-HSI features, and finds texturally advanced and more spectrally similar features for LR-HSI before fusion, which greatly helps to obtain pansharpened HSI with simultaneously high spectral and spatial qualities. Formally, our HyperTransformer consists of four interconnected modules, namely two feature extraction modules for PAN and LR-HSI called FE-PAN and FE-HSI, the attention mechanism, and textural-spectral feature fusion module (TSFF). Our HyperTransformer begins by transforming PAN and LR-HSI to their respective feature space by employing  FE-PAN and FE-HSI, respectively. We then utilize LR-HSI, PAN$\downarrow \uparrow$, and PAN features as queries (Q), keys (K), and values (V) in an attention mechanism to compute texturally advanced and spectrally similar feature representations for LR-HSI features. The computed texturally advanced feature maps  are then mixed with LR-HSI features from a backbone network which constitutes the pansharpened HSI. Furthermore, to obtain visually appealing pansharpened HSIs, we also introduce two new loss terms to the HS pansharpening, namely perceptual loss and transfer-perceptual loss in addition to the widely adopted $L_1$ loss. In summary, this paper makes the following  contributions:
\begin{itemize}[noitemsep]
    \item We propose a novel transformer network called \textit{HyperTransformer} for HS pansharpening which achieves significant improvements over SOTA approaches. To the best of our knowledge, we are one of the first to introduce fusion transformer architecture for HS pansharpening. 
    \item We propose a novel \textit{multi-scale} feature fusion strategy for HS pansharpening which enables our network to effectively capture multi-scale long-range details and cross-feature space dependencies of PAN and LR-HSI by employing HyperTransformers at different scales of the backbone network.
    \item We also introduce two novel loss functions for HS pansharpening, namely \textit{synthesized perceptual loss} and \textit{transfer perceptual loss} which enables our HyperTransformer to learn more powerful feature representations of PAN and LR-HSI. 
\end{itemize}
\vspace{-4mm}
\section{Related Work}
\vspace{-2mm}
\paragraph{Classical approaches.} Classical pansharpening approaches can be divided into four categories: component substitution (CS)~\cite{CS_IHS, CS_PCA, CS_GS}, multi-resolution analysis (MRA)~\cite{MRA_Wavelet, MRA_Lapalacian}, hybrid~\cite{GFPCA}, and Bayesian methods~\cite{Bayesian1, Bayesian3}.  CS-based methods first decompose LR-HSI into spectral and spatial components. Subsequently, the spatial component is substituted with the PAN image and transformed back to the original space by employing the inverse transformation. The widely employed algorithms such as Gram–Schmidt (GS) \cite{GS}, GS-adaptive (GSA)\cite{GSA}, and principal component analysis (PCA) \cite{PCA1,PCA2} are examples of CS. The MRA-based pansharpening methods inject spatial features to LR-HSI by employing a spatial filter. The smoothing filter-based intensity modulation (SFIM) \cite{SFIM},  MTF-GLP (MG) \cite{MTF-GLP}, MTF-GLP with high-pass modulation (MGH) \cite{MTF-GLP-HPM} are examples of MRA. Considering the limitations of the CS and MRA, hybrid methods have been proposed, such as guided filter PCA (GFPCA) \cite{GFPCA}.  Bayesian methods formulate the fusion problem in a Bayesian inference framework. Examples of these include  Bayesian Fusion (BF) \cite{BF}, sparse BF (BSF) \cite{BFS}, and  coupled non-negative matrix factorization (CNMF).

\noindent{\bf{ConvNet-based approaches.}} ConvNet-based pansharpening approaches have recently shown significant progress in pansharpening due to their strong capability to learn high-level features from input data. Among those methods, Masi et al. \cite{pansharpening_by_CNN} are the first  to present a three-layer ConvNet architecture taking the up-sampled LR-HSI staked with PAN as input. Inspired by the wide adaptation of ResNet~\cite{resnet} in image super-resolution tasks, a deep residual pansharpening network (DRPNN) was proposed in \cite{wei2017boosting} to learn the residual image between reference HSI and up-sampled HSI.  Motivated by the 3-D characteristics of HSI data, Palsson et al. \cite{palsson2017multispectral} proposed a 3d-ConvNet which has shown promising results when LR-HSI is corrupted by additive noise. Later, Dian et al. \cite{8295275} proposed a deep pansharpening approach called DHSIS, which integrates priors learned by a ConvNet into the fusion of LR-HSI and PAN features.  In order to improve the spectral prediction capability of HS pansharpening networks, two spectrally predictive ConvNet models called HyperPNN1 and HyperPNN2 were designed in \cite{Hyper-PNN}. More recently, attention mechanisms~\cite{DHP-DARN, ECCV_20_cross_attention,9662053} (i.e., spectral and spatial attention) have been introduced to HS pansharpening to capture long-range details present in PAN and LR-HSI. In~\cite{DHP-DARN} (DHP-DARN),  spectral and spatial attention residual blocks  are utilized to map the residual image between the reference HR-HSI and the upsampled HSI, and has achieved better fusion performance compared to other SOTA methods. However, none of these attention-based methods mentioned above have explored attention for textural-spectral feature fusion process of pansharpening by utilizing feature representations of LR-HSI, PAN$\downarrow \uparrow$, and PAN as queries, keys, and values - which we will explore in this study.
\vspace{-3mm}
\section{Methodology}
    \vspace{-2mm}
    \par
\begin{figure}[tb]
        \centering
        \includegraphics[width=\linewidth]{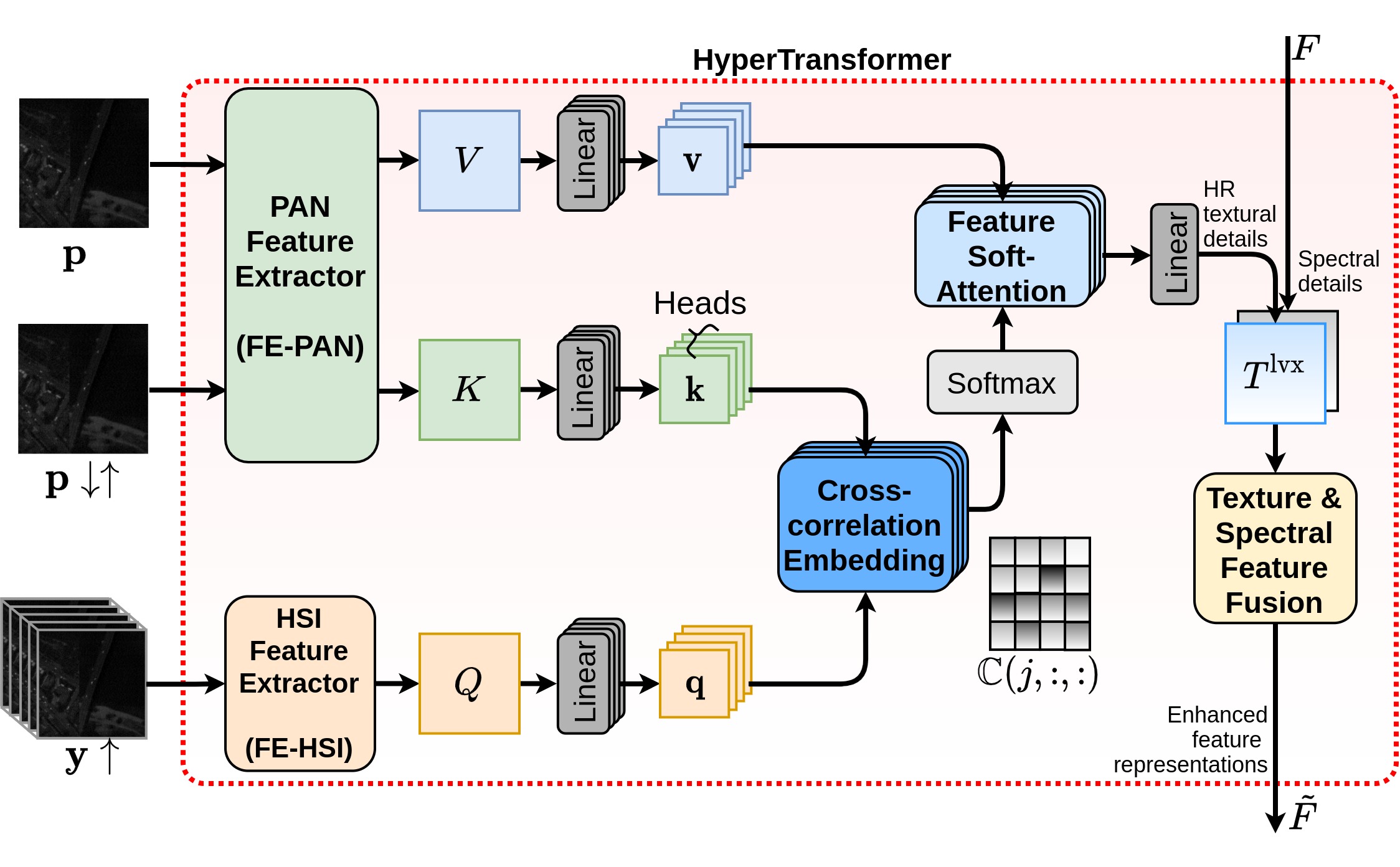}
        \vskip-10pt\caption{Overall structure of the proposed HyperTransformer for textural-spectral feature fusion.}
        \vspace{-5mm}
        \label{fig:hyper_transformer}
    \end{figure}

 The overall structure of our HyperTransformer is shown in Figure \ref{fig:hyper_transformer}, where $\mathbf{p}$, $\mathbf{p}\downarrow \uparrow$, and $\mathbf{y}\uparrow$ represent the PAN image, the sequentially $4\times$ down-sampled and $4\times$ up-sampled PAN image, and  $4\times$ upsampled LR-HSI, respectively. We use bicubic interpolation for upsampling/down-sampling due to its experimentally proven less spatial and spectral distortions for HSIs~\cite{HSI_review_new}. The sequential down-sampling and up-sampling operations make  $\mathbf{p} \downarrow \uparrow$ to domain-consistent with $\mathbf{y}\uparrow$: which is essential for smooth and reliable operation of the attention mechanism.  HyperTransformer takes $\mathbf{p}$, $\mathbf{p}\downarrow \uparrow$, $\mathbf{y}\uparrow$, and the LR-HSI features ($\mathbf{F}$) produced by the backbone as the inputs, and outputs texturally advanced and spectrally similar feature representations of $\mathbf{F}$ (denoted as $\Tilde{\mathbf{F}}$), which will be further used to generate the pansharpened HSI by the backbone network. The proposed transformer has four main modules: two separate feature extractors for PAN and HSIs (FE-PAN and FE-HSI), a feature attention module, and a textural-spectral feature fusion module. Next, we will discuss each of the modules in detail.  
    
    \vspace{-2mm}
    \subsection{Feature Extractors for PAN and LR-HSI} 
    \label{feature_extractors}
    \vspace{-2mm}
    \par
\begin{figure*}[tb]
        \centering
        \includegraphics[width=0.95\linewidth]{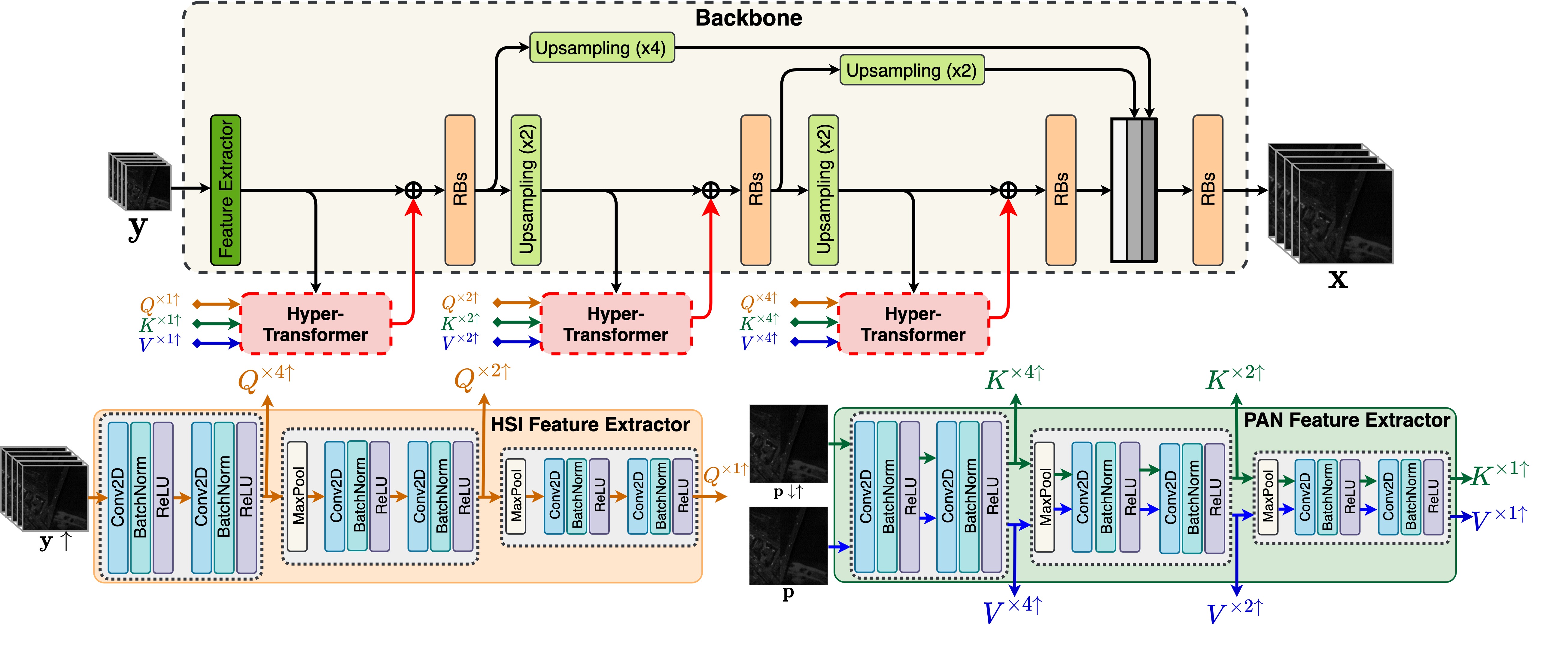}
        \vskip-10pt \caption{The complete pansharperning network. Note that we apply  HyperTransformer at three scales: $\times1 \uparrow$, $\times2 \uparrow$, and $\times4 \uparrow$. RBs denotes the residual blocks.\vspace{-2mm}}
        \vspace{-4mm}
        \label{fig:complete_network}
    \end{figure*}

 We design two separate Feature Extractors (FE) to obtain HR textural and spectral features from PAN and LR-HSI, respectively. We employ VGG-like network architecture for the FEs (see Figure \ref{fig:complete_network}). The VGG-like design encourages the learning of precise mutual spectral and textural information in the LR-HSI and PAN image. The outputs from the FEs define the Query $(Q)$, Key $(K)$, and Value $(V)$ features, which are the three basic elements of the attention mechanism inside the HyperTransformer. Formally, $Q$, $K$, and $V$ are obtained  as follows:
    \setlength{\belowdisplayskip}{0pt} \setlength{\belowdisplayshortskip}{0pt}
    \setlength{\abovedisplayskip}{0pt} \setlength{\abovedisplayshortskip}{0pt}
    \begin{align}
        Q &= f_{\text{FE-HSI}}(\mathbf{y} \uparrow),\\
        K &= f_{\text{FE-PAN}}(\mathbf{p} \downarrow \uparrow),\\
        V &= f_{\text{FE-PAN}}(\mathbf{p}),
    \end{align}
    where $f_{\text{FE-HSI}}(\cdot)$ and $f_{\text{FE-PAN}}(\cdot)$ are the parametric representations of FE-HSI and FE-PAN, respectively. 
    
    \vspace{-2mm}
    \subsection{HR Texture Transfer through Multi-Head Feature Soft-Attention (MHFSA)}
    \vspace{-2mm}
    Feature attention aims to identify spectrally similar and texturally superior feature representations for LR-HSI features, which will be further used to produce pansharpened HSI by the backbone network. For this purpose, we utilize a multi-head feature soft-attention (MHFSA) mechanism instead of a single-head feature-attention due to experimentally recognized better spatial and spectral properties of pansharpened HSI. To facilitate MHFSA, we first derive a set of $N$ global descriptors for each feature in $Q$, $K$, and $V$ by utilizing $N$ fully-connected layers. Next, we compute feature soft-attention for each descriptor in parallel. Finally, we concatenate the output feature descriptors from feature soft-attention and employ linear layers to convert them back to the original feature space. The proposed MHFSA mechanism greatly assists the network in extracting cross-feature space dependencies between PAN and LR-HSI and long-range details of PAN.
    \vspace{-0.5cm}
    \paragraph{Obtaining a set of $N$ global feature descriptors.} We first reshape queries $Q \in \mathbb{R}^{f_q \times w \times h}$, keys $K \in \mathbb{R}^{f_k \times w \times h}$, and values $V \in \mathbb{R}^{f_v \times w \times h}$ into 2D tensors $q \in \mathbb{R}^{f_q \times wh}$, $k \in \mathbb{R}^{f_k \times wh}$, and $v \in \mathbb{R}^{f_v \times wh}$. Note that we discard the batch dimension from our notations for simplicity. The $f_q$, $f_k$, and $f_v$ represent the number of feature maps in $Q$, $K$, and $V$, and $w$ and $h$ represent the width and height of a feature map, respectively. Next, we utilize a set of $N$ linear (fully-connected) layers  to transform each feature map in $q$, $k$, and $v$ into a set of $N$ global descriptors (heads) to facilitate multi-head feature soft-attention. The resulting $N$ global descriptors for each feature map of $q$, $k$, and $v$ represent  3-D tensors $\mathbf{q} \in \mathbb{R}^{f_q, N, \beta wh}$, $\mathbf{k} \in \mathbb{R}^{f_q, N, \beta wh}$, and $\mathbf{v} \in \mathbb{R}^{f_q, N, \beta wh}$, where $\beta$ denotes the dimensionality reduction ratio. Formally, we can define the above process as:
    \begin{align}
        \mathbf{q}(j, i,:) &= f_\text{linear-q}^i (q(j, :)),\\
        \mathbf{k}(j, i,:) &= f_\text{linear-k}^i (k(j, :)),\\
        \mathbf{v}(j, i,:) &= f_\text{linear-v}^i (v(j, :)),
    \end{align}
    where $f_\text{linear-q}^i (\cdot)$, $f_\text{linear-k}^i (\cdot)$, and $f_\text{linear-v}^i (\cdot)$ are the parametric representation of the $i$-th linear-layer associated with query, key, and value, $q(j, :)$, $k(j, :)$, and $v(j, :)$ are the 1-D representation of the $j$-th feature map of $q$, $k$, and $v$, and $\mathbf{q}(j, i, :)$, $\mathbf{k}(j, i, :)$, and $\mathbf{v}(j, i, :)$ are the $i$-th global descriptor of the $j$-th feature map of $q$, $k$, and $v$, respectively.
    \vspace{-0.5cm}
    \paragraph{Feature Cross-Correlation Embedding (FCCE).} We compute the feature cross-correlation matrices  between query ($\mathbf{q}$) and key ($\mathbf{k}$) for all $N$ descriptors separately, and represent them in a 3-D matrix $\mathbb{C} \in \mathbb{R}^{N \times f_q \times f_k}$. In order to efficiently compute feature cross-correlation for all the $N$ descriptors in parallel using matrix multiplication (i.e., without any ``for" loops), we first permute the  first two dimensions (dim 0 and 1) of $\mathbf{q}$, $\mathbf{k}$, and $\mathbf{v}$. The resulting permuted matrices are denoted as $\mathbf{q}' \in \mathbb{R}^{N \times f_q \times \beta wh}$, $\mathbf{k}' \in \mathbb{R}^{N \times f_k \times \beta wh}$, and $\mathbf{v}' \in \mathbb{R}^{N \times f_v \times \beta wh}$. We then compute the feature cross-correlation for $N$ descriptors at once as follows: 
    \begin{equation}
        \mathbb{C} = \text{MatMul}((\mathbf{q}' - mean(\mathbf{q}')), (\mathbf{k}' - mean(\mathbf{k}'))^T),
    \end{equation}
    where $T$ denotes the matrix transpose operation, MatMul denotes the batch matrix multiplication on dim-1 and 2, and $mean(\cdot)$ denotes the mean value. The rows of cross-correlation matrix for the $j$-th descriptor (i.e., $\mathbb{C}(j,:,:)$) tells us how a given query descriptor (i.e., an LR-HSI feature) correlates with all the key descriptors. In other words, it extracts the cross-feature space dependencies between LR-HSI features and PAN features (note that the queries and values are the feature representations of LR-HSI and PAN$\downarrow \uparrow$, respectively). We then utilize a Softmax layer along the rows of each correlation matrix in $\mathbb{C}$ to get the row-normalized cross-correlation matrices. Formally, we can define this process as follows:
    \begin{align}
        \Tilde{\mathbb{C}} &= \text{Softmax}(\mathbb{C}, \text{dim} = 1),
    \end{align}
    where $\Tilde{\mathbb{C}}$ contained row normalized (sums to 1) feature cross-correlation matrices of $N$ descriptors.
    \vspace{-0.5cm}
    \paragraph{Multi-Head Feature Soft-Attention (MHFSA).} 
    We then compute feature soft-attention on the $N$ descriptors at once using matrix multiplication as follows:
    \begin{align}
        t & = \text{MatMul} (\Tilde{\mathbb{C}}, \mathbf{v}')
    \end{align}
    where $t \in \mathbb{R}^{N, f_q, \beta hw}$ is the output of the MHFSA. Next, we permute the dimensions of $t$ to its original format ($\mathbb{R}^{N, f_q, \beta hw} \rightarrow \mathbb{R}^{f_q, N, \beta hw}$) and apply a linear layer followed by reshaping to obtain texturally advanced feature representation $T \in \mathbb{R}^{f_q, \times h \times w}$ from  HyperTransformer as follows:
    \begin{align}
        T \in \mathbb{R}^{f_t \times hw} &= \text{Linear} (t),\\
        T \in \mathbb{R}^{f_t \times h \times w} &\leftarrow \text{Reshape} (T),
    \end{align}
    where $f_t$ is the number of features in $T$. Note that to comply with matrix multiplications and additions we set $f_q = f_k = f_v = f_t$.
    \vspace{-2mm}
    \subsection{Textural-Spectral Feature Fusion (TSFF)}
    \vspace{-2mm}
    The texturally advanced and spectrally similar feature representation $T$ of LR-HSI features $Q$ that we obtain through MHFSA are further concatenated with spectral features $F$ from the backbone network as shown in Figure \ref{fig:hyper_transformer}. Next, we employ a $3 \times 3$ convolution followed by a Batch Normalization (BN) layer to fuse textural-spectral details together and generate the residual component required for the backbone network which will be further used to generate the pansharpened HSI. Formally, the process inside TSFF can be define as follows:
    \begin{equation}
        \Tilde{F} = \text{BatchNorm}(\text{Conv}( \text{Cat} (T, F))),
    \end{equation}
    where $\Tilde{F}$ is the output of  HyperTransformer which will be further used by the backbone network to generate pansharpened HSI, and Cat denotes concatenation operation.
    \vspace{-2mm}
    \subsection{Multi-Scale Feature Fusion (MSFF)}
    \vspace{-2mm}
    \par Unlike the conventional pansharpening methods that fuse textural features from PAN only at the HR scale, we inject textural details from our HyperTransformer to the backbone network at multiple spatial scales, as depicted in Figure \ref{fig:complete_network}. In particular, we inject HR-textural details at three spatial scales: (1) LR-HSI spatial scale (denoted by $\times 1 \uparrow$), (2) two times upsample LR-HSI spatial scale (denoted by $\times 2 \uparrow$), and (3) desired HR spatial scale (denoted by $\times 4 \uparrow$). Accordingly, we denote the inputs and outputs of  HyperTransformer as $X^{\times s \uparrow}$ in general where $X$ could be $Q, K, V, F$ or $\Tilde{F}$, and $s$ represents the spatial-scale: $1$, $2$, or $4$ (see Figure \ref{fig:complete_network}). Injecting HR-textural knowledge at multiple spatial scales helps the network to capture multi-scale long-range details and multi-scale cross-feature space dependencies of PAN and LR-HSI, resulting in better spatial and spectral quality of pansharpened HSI.
    
    \vspace{-2mm}
    \subsection{Loss Functions}
    \vspace{-2mm}
   We utilize a combination of three loss functions to train our network. 
    \vspace{-0.6cm}
    \paragraph{Reconstruction loss.} We use $L_1$ loss as the reconstruction loss:
    \begin{equation}
        \mathcal{L}_{\text{rec}} = \frac{1}{CWH} \norm{\mathbf{x}_{\text{ref}} - \mathbf{x}}_1,
    \end{equation}
    where $\mathbf{x}_{\text{ref}}$ is the target HR-HSI, $\mathbf{x}$ is the predicted HR-HSI, and $(C, H, W)$ is the size of the HR-HSI. We utilize $L_1$ loss which has been demonstrated to perform better compared to $L_2$ loss for HS pansharpening \cite{HSI_review_new}.
    \vspace{-0.6cm}
    \paragraph{VGG perceptual loss.} The VGG perceptual loss has been originally demonstrated useful for RGB super-resolution tasks to enhance the visual quality of images~\cite{johnson2016perceptual}. The underlying idea of the perceptual loss is to enhance the similarity in the feature space between the predicted image and the target image. The feature maps of predicted and target image are obtained from a pre-trained VGG network which is trained on RGB images. In order to evaluate VGG loss for HSI images, we first synthesize RGB image for a given HSI by defining Gaussian approximated spectral response curve for R, G, and B bands. Next, we evaluate the perceptual loss as:
    \begin{equation}
        \mathcal{L}_{\text{vgg-per}} = \frac{1}{C_i W_i H_i} \norm{f_i^{\text{vgg}}(\mathbf{x}_{\text{ref}}^{\text{rgb}}) - f_i^{\text{vgg}}(\mathbf{x}^{\text{rgb}})}_2,
    \end{equation}
    where $f_i^{\text{vgg}}(\cdot)$ denotes the $i$-th layer’s feature map of VGG-19~\cite{simonyan2014very}, and $(C_i,H_i,W_i)$ represents the shape of the feature map at that layer. $\mathbf{x}_{\text{ref}}^{\text{rgb}}$ and $\mathbf{x}^{\text{rgb}}$ are the synthesized RGB images of the target HSI $\mathbf{x}_{\text{ref}}$ and predicted HSI $\mathbf{x}$, respectively.
    \vspace{-0.7cm}
    \paragraph{Transfer perceptual loss.} The transfer perceptual loss constraints the predicted HSI image $\mathbf{x}$ to have similar texture features to the transferred texture features $T$ from HyperTransformer, which makes our HR texture transfer process more effective. The  transfer perceptual loss is calculated as follows:
    \begin{equation}
        \mathcal{L}_{\text{t-per}} = \frac{1}{C_s W_s H_s} \norm{f_{\text{FE-HSI}}(\mathbf{x})^s - T^s}_2,
    \end{equation}
    where $T^s$ denotes the transferred feature map at the $s$-th spatial scale (i.e., 1, 2, or 4), $f_{\text{FE-HSI}}(\mathbf{x})^s$ is the feature map at the $s$-th spatial scale from the HSI feature extractor, and $(C_s, W_s, H_s)$ represents the size of the feature map at that scale, respectively. 
    
    The overall loss function we use to train our network is defined as follows:
    \begin{equation}
        \mathcal{L}_{\text{overall}} = \lambda_{\text{rec}} \mathcal{L}_{\text{rec}} + \lambda_{\text{vgg-per}} \mathcal{L}_{\text{vgg-per}} + \lambda_{\text{t-per}} \mathcal{L}_{\text{t-per}},
    \end{equation}
    where $\lambda_{\text{rec}}, \lambda_{\text{vgg-per}}$ and $\lambda_{\text{t-per}}$ are regularization constants. We set $\lambda_{\text{rec}}= 1.0, \lambda_{\text{vgg-per}}=0.1$ and $\lambda_{\text{t-per}}=0.05$.
\vspace{-2mm}
\section{Experiments}
    \vspace{-2mm}
    \subsection{Datasets and Performance Metrics} 
    \vspace{-2mm}
    We use three publicly available and widely used HSI datasets for our experiments, namely Pavia Center~\cite{pavia_center_dataset}, Botswana~\cite{botswana_dataset}, and Chikusei~\cite{chikusei_data}. Following the experimental and data preparation procedure outlined in \cite{DIP-HyperKite} and \cite{DHP-DARN}, we create cubic patches of size $102 \times 160 \times 160$, $145 \times 120 \times 120$, and $128 \times 256 \times 256$ as the reference HSIs $(\mathbf{x}_{\text{\text{ref}}})$ for the Pavia Center, Botswana, and Chikusei datasets, respectively. We then utilize Wald’s protocol~\cite{walds_protocol1, walds_protocol2} to generate PAN and LR-HSI from the reference HSIs. As part of the Wald's protocol, we use $8 \times 8$ Gaussian filter followed by down-sampling operator with scaling factor of $4$ to generate LR-HSI images for all the three datasets. We randomly select $\sim 80\%$ of cubic patches to form the training set for each dataset and the rest of the cubic patches are used to form the testing set. We use  10-th, 10-th, and 12-th spectral bands as the \textcolor{blue}{\bf blue}-band,  30-th, 35-th, and 20-th spectral bands as the \textcolor{green}{\bf green}-band, and 60-th, 61-th, and 29-th spectral bands as the \textcolor{red}{\bf red}-band when synthesizing the RGB image for Pavia Center, Botswana, and Chikusei datasets, respectively.
    \par To evaluate the quality of the proposed pansharpening method, we use different spatial and spectral quality measures. Following \cite{HSI_review_new, DHP-DARN, DIP-HyperKite}, we use Cross-Correlation (CC), Spectral Angle Mapping (SAM), Root Mean Square Error (RMSE), Reconstruction Signal to Noise Ratio (RSNR), Errur Relative Globale Adimensionnelle Desynthese (ERGAS), and Peak Signal to Noise Ratio (PSNR). These measures have been widely used in the HSI processing community and are appropriate for evaluating fusion in spectral and spatial resolutions.
    
    \begin{table*}[!htbp]
    	\begin{minipage}{\linewidth}
    		\caption{The average quantitative pansharpening results on the Pavia Center~\cite{pavia_center_dataset}, Botswana~\cite{botswana_dataset}, and Chikusei dataset~\cite{Chikusei_dataset}.$^{*}$}
    		\label{tab:main_quantitative-results}
    		\centering
    	\resizebox{\textwidth}{!}{%
    	\small 
	    \begin{tabular}{lp{0.5cm}p{0.6cm}p{0.6cm}p{0.7cm}p{0.5cm}p{0.01cm}p{0.5cm}p{0.6cm}p{0.6cm}p{0.7cm}p{0.5cm}p{0.01cm}p{0.5cm}p{0.6cm}p{0.6cm}p{0.7cm}p{0.4cm}}
		\toprule
		\multirow{2}{*}{\parbox[c]{.2\linewidth}{\centering Method}} & \multicolumn{5}{c}{Pavia Center Dataset~\cite{pavia_center_dataset}} & & \multicolumn{5}{c}{Botswana Dataset~\cite{botswana_dataset}} & & \multicolumn{5}{c}{Chikusei Dataset~\cite{Chikusei_dataset}} \\ 
		\cmidrule{2-6} \cmidrule{8-12} \cmidrule{14-18}
		            &{CC}&{SAM}&{RMSE}&{ERGAS}&{PSNR}&&     {CC}&{SAM}&{RMSE}&{ERGAS}&{PSNR}&&     {CC}&{SAM}&{RMSE}&{ERGAS}&{PSNR}\\
		             & & &$\times 10^{-2}$& & && & &$\times 10^{-2}$ & &  &&  & &$\times 10^{-2}$ &  & \\
		\midrule
		\textbf{PCA~\cite{PCA1}} \textsuperscript{PERS-2014}		& 
		0.845 & 8.92 & 3.45 &6.64 & 31.26&&                
		0.943 & 2.38 & 1.98 &2.22 & 40.03&&
		0.887 & 6.99 & 2.47 &7.71 & 30.98\\
		
		\textbf{GFPCA~\cite{GFPCA}} \textsuperscript{DataFusion-2014}	& 
		0.902 & 8.31 & 3.98 &7.44 & 29.09&&                
		0.901 & 2.66 & 2.45 &2.75 & 37.83&&
		0.883 & 4.76 & 1.98 &7.00 & 30.96\\
		
		\textbf{BF~\cite{BF}} \textsuperscript{JSTSP-2015} & 
		0.918 & 9.60 & 3.44 &6.63 & 30.22&&                
		0.931 & 2.47 & 1.88 &2.34 & 40.01&&
		0.903 & 5.15 & 1.94 &6.62 & 37.89\\
		
		\textbf{BFS~\cite{BFS}} 	\textsuperscript{TGRS-2015}	& 
		0.925 & 8.10 & 3.05 &6.00 & 31.09&&                
		0.932 & 2.39 & 1.85 &2.32 & 40.15&&
		0.917 & 4.69 & 1.72 &6.39 & 37.99\\
		
		\textbf{SFIM~\cite{SFIM}} \textsuperscript{IJRS-2000}		& 
		0.946 & 6.76 & 2.55 &5.43 & 32.61&&                
		0.932 & 3.44 & 2.81 &2.25 & 39.58&&
		0.928 & 3.79 & 1.43 &6.43 & 39.55\\
		
		\textbf{GS~\cite{GS}} \textsuperscript{TGRS-2007}		& 
		0.961 & 6.62 & 2.55 & 4.95 & 32.93&&                
		0.946 & 2.34 & 1.93 &2.17 & 40.14&&
		0.733 & 5.64 & 2.96 &8.17 & 35.13\\
		
		\textbf{GSA~\cite{GSA}} 	\textsuperscript{US Pat.-2000}	& 
		0.950 & 7.15 & 2.34 &4.70 & 33.52&&                
		0.955 & 2.04 & 1.59 &1.85 & 41.89&&
		0.943 & 3.52 & 1.42 &4.30 & 41.38\\
		
		\textbf{MGH~\cite{MTF-GLP-HPM}} 	\textsuperscript{PERS-2006}	& 
		0.955 & 6.81 & 2.25 &4.77 & 33.97&&                
		0.960 & 2.07 & 1.54 &1.69 & 42.43&&
		0.929 & 3.82 & 1.45 &6.40 & 39.85\\
		
		\textbf{CNMF~\cite{CNMF}} \textsuperscript{TGRS-2011}		& 
		0.960 & 6.64 & 2.20 &4.39 & 34.14&&                
		0.942 & 2.61 & 1.73 &2.10 & 40.98&&
		0.900 & 4.72 & 1.91 &5.75 & 39.65\\
		
		\textbf{MG~\cite{MTF-GLP}} \textsuperscript{TGRS-2002}  		& 
		0.956 & 6.55 & 2.20 &4.45 & 34.12&&                
		0.960 & 2.02 & 1.51 &1.68 & 42.47&&
		0.938 & 3.81 & 1.52 &4.41 & 41.05\\
		
		\textbf{HySure~\cite{hysure}} \textsuperscript{TGRS-2014} 		& 
		0.966 & 6.13 & 1.80 &3.77 & 35.91&&                
		0.956 & 2.15 & 1.46 &1.77 & 42.30&&
		0.960 & 2.98 & 1.13 & \bf 3.69 & 43.14\\
		              
		\textbf{HyperPNN~\cite{Hyper-PNN}} \textsuperscript{JST-RS-2019} 	& 
		0.967 & 6.09 & 1.67 &3.82 & 36.70   &&
		0.970 & 1.67 & 1.15 &1.44 & \bf 44.45 &&
		0.946 & 3.97 & 1.11 &4.77 & 41.57\\
		              
		\textbf{PanNet~\cite{pannet_iccv17}} \textsuperscript{ICCV-2017}	& 
		0.968 & 6.36 & 1.83 & 3.89 & 35.61 &&   
		0.926 & 2.17 & 1.53 & 2.82 & 40.41 &&
		0.956 & 3.79 & \bf 0.88 & 5.32 & 41.90\\
		
		\textbf{DARN~\cite{DHP-DARN}} \textsuperscript{TGRS-2020}		& 
		\bf 0.969 & 6.43 & \bf 1.56 & \bf 3.95 & \bf 37.30&&                
		\bf 0.973 & \bf 1.58 & \bf 1.09 & \bf 1.35 & 44.42&&
		0.953 & 3.60 & 1.05 &4.44 & 42.24\\
		              
		\textbf{HyperKite~\cite{DIP-HyperKite}} \textsuperscript{TGRS-2021} 	& 
		\textcolor{blue}{\bf 0.980} & \bf 5.61 & \textcolor{blue}{\bf 1.29} & \textcolor{blue}{\bf 2.85} & \textcolor{blue}{\bf 38.65}&&                
		\textcolor{blue}{\bf 0.979} & \textcolor{blue}{\bf 1.46} & \textcolor{blue}{\bf 1.01} & \textcolor{blue}{\bf 1.21} & \textcolor{blue}{\bf 45.53}&&
		\textcolor{blue}{\bf 0.974} & \bf 2.85 & 1.03 & \textcolor{blue}{\bf 3.62} & \bf 43.53\\
		              
		\textbf{SIPSA~\cite{2021_pan_shift_inv}} \textsuperscript{CVPR-2021} 	& 
		0.948 & \textcolor{blue}{\bf 5.27} & 2.38 & 4.52 & 33.65 &&    
		0.901 & 2.34 & 2.20 & 2.54 & 38.55 &&
		0.947 & 2.87 & 1.06 & 5.09 & 41.02 \\
		
		\textbf{GPPNN~\cite{2021_CVPR}} \textsuperscript{CVPR-2021} &   
		0.963 & 6.52 & 1.91 & 4.05 &  35.36 &&
		0.962 & 1.90 & 1.36 & 1.65 & 43.01 &&
		\bf 0.970 & \textcolor{blue}{\bf 2.75} & \textcolor{blue}{\bf 0.66} & 4.24 & \textcolor{blue}{\bf 44.07} \\
		
		\bf Ours  & \textcolor{red}{\textbf{0.989}} & \textcolor{red}{\textbf{3.85}} & \textcolor{red}{\textbf{0.87}} & \textcolor{red}{\textbf{2.01}} & \textcolor{red}{\textbf{43.80}} && \textcolor{red}{\textbf{0.982}} & \textcolor{red}{\textbf{1.18}} & \textcolor{red}{\textbf{0.89}} & \textcolor{red}{\textbf{1.04}} & \textcolor{red}{\textbf{46.97}} && \textcolor{red}{\textbf{0.980}} & \textcolor{red}{\textbf{2.40}} & \textcolor{red}{\textbf{0.57}} & \textcolor{red}{\textbf{3.12}} & \textcolor{red}{\textbf{45.87}}\\
		\bottomrule
		\multicolumn{18}{p{19.1cm}}{\footnotesize{$^{*}$Higher values of CC and PSNR, and lower values of SAM, RMSE, and ERGAS indicate good pansharpening performance. The ideal values of CC, SAM, RMSE, ERGAS, and PSNR are $1$, $0$, $0$, $0$, and $\infty$, respectively. Color convention: \textcolor{red}{\bf best}, \textcolor{blue}{\bf 2nd-best}, and \textcolor{black}{\bf 3rd-best}.%
		}}
	\end{tabular}}
    	\end{minipage}\\
    	\begin{minipage}{\linewidth}
    		\centering
    		\vspace{1mm}
    		\includegraphics[width=\linewidth]{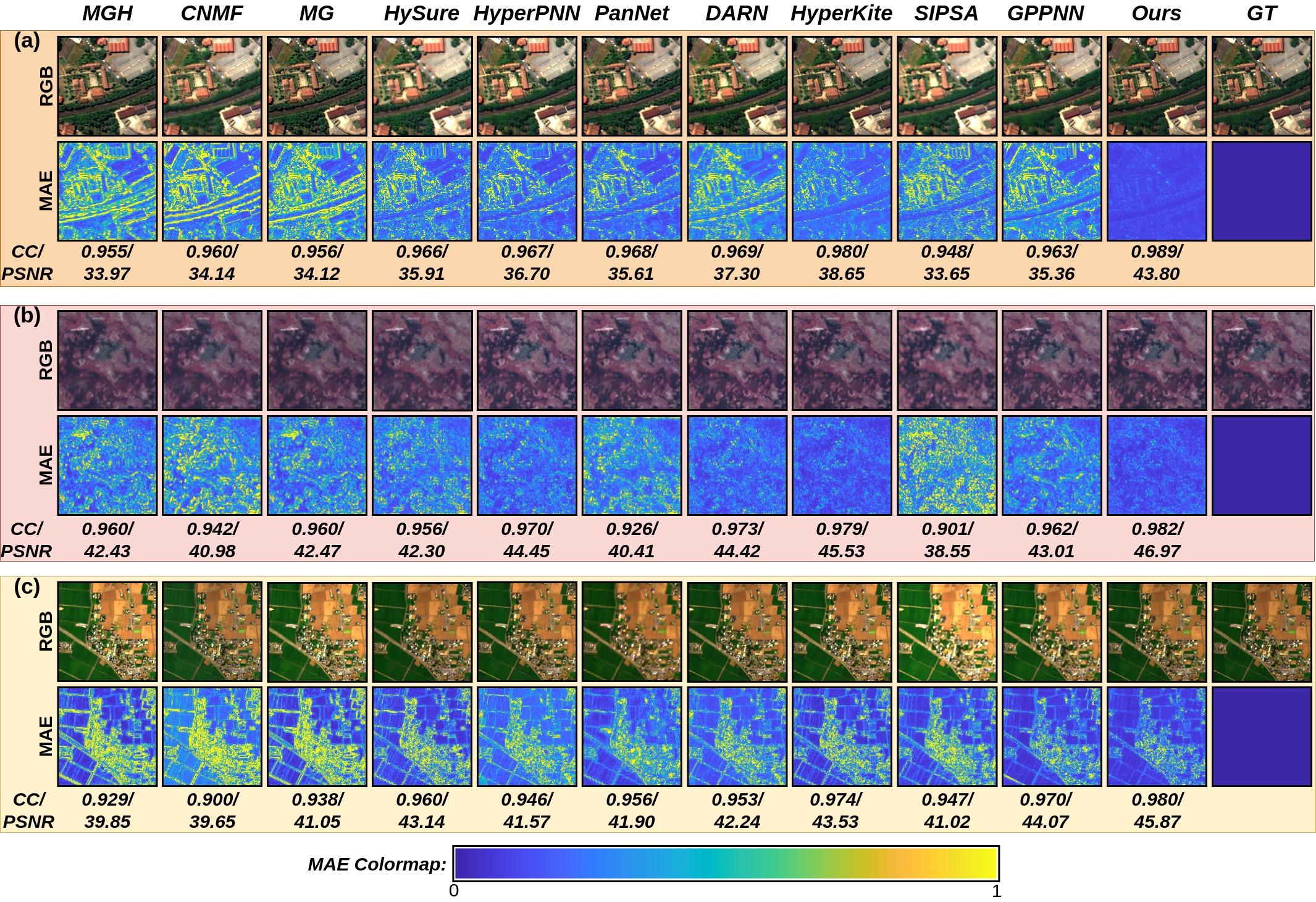}
             \vskip-10pt\caption{Visual results generated by different pansharpening algorithms (left to right: MGH~\cite{MTF-GLP-HPM}, CNMF~\cite{CNMF}, MG~\cite{MTF-GLP}, HySure~\cite{hysure}, HyperPNN~\cite{Hyper-PNN}, PanNet~\cite{pannet_iccv17}, DARN~\cite{DHP-DARN}, HyperKite~\cite{DIP-HyperKite}, SIPSA~\cite{cvpr_2021_sipsa}, GPPNN~\cite{2021_CVPR}, HyperTransformer (\textbf{ours}), and Ground Truth (GT)) for \textbf{(a)} Pavia Center~\cite{pavia_center_dataset} (20-th patch), \textbf{(b)} Botswana~\cite{botswana_dataset} (12-th patch), \textbf{(c)} Chikusei datasets~\cite{Chikusei_dataset} (32-nd patch). MAE denotes the (normalized) Mean Absolute Error across all spectral bands.}
            \label{fig:main_qualitative_results}
    	\end{minipage}
    \end{table*}
    \vspace{-1mm}
    \subsection{Results and Discussion}
    \vspace{-2mm}
    To demonstrate the effectiveness of HyperTransformer, we compare our model with both classical and  ConvNet-based SOTA methods. The classical methods include PCA\cite{PCA1}, GFPCA\cite{GFPCA}, BF\cite{BF}, BFS\cite{BFS}, SFIM\cite{SFIM}, GS\cite{GS}, GSA\cite{GSA}, MGH\cite{MTF-GLP-HPM}, CNMF\cite{CNMF}, MG\cite{MTF-GLP}, and HySure\cite{hysure},  among which HySure has achieved the SOTA performance on CC, SAM, RMSE, ERGAS, and PSNR in recent years. As for the ConvNet-based methods, HyperPNN \cite{Hyper-PNN}, PanNet \cite{pannet_iccv17}, DHP-DARN (abbreviated as DARN) \cite{DHP-DARN}, DIP-HyperKite (abbreviated as HyperKite) \cite{DIP-HyperKite}, SIPSA~\cite{cvpr_2021_sipsa}, and GPPNN~\cite{2021_CVPR} are six recent SOTA methods which significantly outperform previous ConvNet-based methods.
    \vspace{-0.5cm}
    \paragraph{Quantitative results.} Table \ref{tab:main_quantitative-results} shows the quantitative evaluation results. As shown in the table, our HyperTransformer significantly outperforms both classical and ConvNet-based SOTA methods on all three datasets. The percentage improvement in CC/SAM/RMSE/ERGAS/PSNR performance measures for Pavia Center, Botswana, and Chikusei datasets are $\sim$0.9/26.9/32.6/29.4/13.3\%, $\sim$0.3/19.2/11.9/14.0/3.2\%, and $\sim$0.6/12.7/13.6/13.8/4.1\%, respectively. These quantitative comparison results demonstrate the superiority of  HyperTransformer over the SOTA approaches.
    \vspace{-0.5cm}
    \paragraph{Qualitative results.} Figure \ref{fig:main_qualitative_results} shows the visual evaluation results where we randomly select one image from the testing set of each dataset and present the synthesized RGB images of HSIs along with the corresponding mean absolute error (MAE) images between the reconstructed HSIs and the reference HSI.  Though the difference in the synthesized RGB images is minute, we can observe the difference between each method from the MAE images. As can be seen from the MAE images, our HyperTransformer achieves significantly lower MAE than all the other methods. These visual results further demonstrate the excellent ability of HyperTransformer to extract fine details more effectively.
    \vspace{-1mm}
    \subsection{Ablation Studies}
    \begin{figure}[tb]
        \centering
        \includegraphics[width=0.8\linewidth]{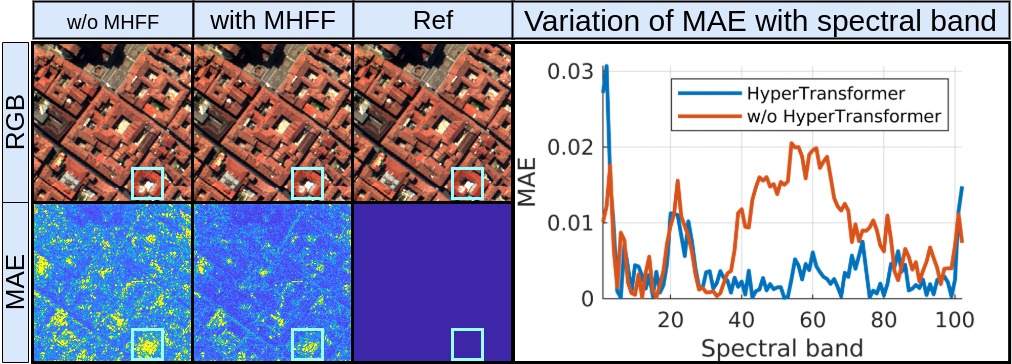}
        \vskip-8pt \caption{The visual results for the ablation study to demonstrate the effect of HyperTransformer for HS pansharpening on the Pavia Center dataset.}
        \label{fig:ablation_transformer}
    \end{figure}
    \begin{table}[tb]
        \centering
         \caption{Ablation study on the number of heads ($N$) in HyperTransformer for the Pavia Center dataset.\vspace{-3mm}}
        \begin{tabular}{c|lllll}
            \hline
            \hline
            $N$  &  CC & SAM & RMSE  & ERGAS & PSNR\\
            \hline
            \hline
            B/L     & 0.981 & 4.88 & 0.0133     & 2.84     & 38.71\\
            \hline
            1       & 0.975 & 4.32  & 0.0103    & 2.31      & 40.59\\
            2       & 0.976 & 4.19  & 0.0095    & 2.18      & 42.52\\
            8       & \bf 0.987 & 4.06  & 0.0092    & 2.13      & \bf 43.20\\
            \textbf{16} & \textcolor{red}{\bf 0.989} & \textcolor{red}{\bf 3.85} & \textcolor{red}{\bf 0.0087} & \textcolor{red}{\bf 2.01} & \textcolor{red}{\bf 43.80}\\
            32& \textcolor{blue}{\bf 0.988} & \textcolor{blue}{\bf 4.02} & \textcolor{blue}{\bf 0.0090} & \textcolor{blue}{\bf 2.10} & \textcolor{blue}{\bf 43.47}\\
            64& \bf 0.987 & \bf 4.04 & \bf 0.0091 & \bf 2.12 & 43.19\\
            \hline
            \hline
        \end{tabular}
        \label{tab:ablation_multiheads}
    \end{table}
    
    \vspace{-2mm}
    \paragraph{HyperTransformer.} To further demonstrate the effectiveness of HyperTransformer on HS pansharpening, we conduct an ablation study, and results are presented in Table \ref{tab:ablation_multiheads} and Figure \ref{fig:ablation_transformer}. In this study, we consider the baseline results (B/L) as the results from our proposed pansharpening network without feature attention mechanism (i.e., we bypass the MHFSA and consider transferred texture features $(T$) as PAN features $(V)$). The proposed HyperTransformer significantly improves the baseline results in CC/SAM/RMSE/ERGAS/PSNR by $\sim$1.5/21/35/29/13\% when $N=16$, respectively. In addition, Figure \ref{fig:ablation_transformer} depicts the synthesized RGB images, MAE plots, and variation of average MAE with spectral bands for a randomly selected region (marked by blue color) for with and without MHFSA. From these plots also we can clearly observe the reduction in MAE over the spectral bands (specially in infrared region) when we utilize HyperTransformer.  All of these qualitative and quantitative comparisons empirically show the effectiveness of our HyperTransformer which captures long-range and cross-feature space dependencies of PAN and LR-HSI for HS pansharpening.
    \vspace{-0.5cm}
    \begin{figure}[tb]
        \centering
        \includegraphics[width=\linewidth]{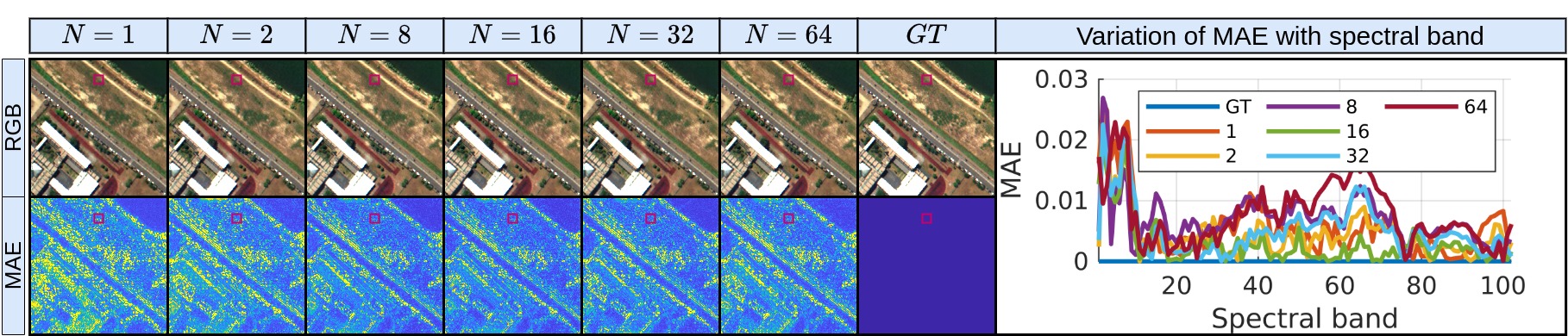}
        \vskip-10pt \caption{The visual results for the ablation study on the number of heads ($N$) in our HyperTransformer. \vspace{-5mm}}
        
        \label{fig:ablation_multihead}
    \end{figure}

    \paragraph{Number of global descriptors $N$.} Table \ref{tab:ablation_multiheads} and Figure \ref{fig:ablation_multihead} present the results when we increase the number of global descriptors in our HyperTransformer from 1 to 64. We can see that increasing the number of global descriptors results in significant improvement over spatial and spectral performance measures. However, after $N=16$, the performance metrics start getting saturated or start degrading. Therefore, we set $N=16$ as the optimal value. We can further verify these quantitative results by observing the qualitative results shown in Figure \ref{fig:ablation_multihead} in which we observe overall low MAE across all spectral bands when $N=16$. Therefore, this study clarifies the reason for selecting $N=16$.
    \vspace{-0.5cm}
    \begin{table}[tb]
        \centering
        \caption{Ablation study on utilizing HyperTransformer at multiple scales for the Pavia Center dataset.\vspace{-3mm}}
        \begin{tabular}{p{0.05\linewidth}p{0.05\linewidth}p{0.05\linewidth}|p{0.07\linewidth}p{0.06\linewidth}p{0.1\linewidth}p{0.11\linewidth}p{0.1\linewidth}}
            \hline
            \hline
            $\times1$  &  $\times2$ & $\times4$ & CC & SAM & RMSE  & ERGAS & PSNR\\
            \hline
            \hline
            \multicolumn{3}{c|}{B/L} & 0.956 & 4.86     & 0.0204     & 3.90 & 35.81\\
            \hline
            \cmark    & \xmark           & \xmark           & 0.975 & 4.76 & 0.0149 & 3.00 & 38.42\\
            \xmark             & \cmark  & \xmark           & 0.985 & 4.29 & 0.0108& 2.40 & 40.80\\
            \cmark    & \cmark  & \xmark           & 0.985 & 4.41 & 0.0109& 2.38 & 41.02\\
            \xmark             & \xmark           & \cmark  & \bf 0.986 & \bf 4.01 & 0.0098 & 2.21 &42.88\\
            \cmark    & \xmark           & \cmark  & \textcolor{blue}{\bf 0.988} & \textcolor{blue}{\bf 3.96} & \textcolor{black}{\bf 0.0092} & \bf 2.20 & \bf 43.58\\
            \xmark             & \cmark  & \cmark  & \textcolor{blue}{\bf 0.988} & \textcolor{red}{\bf 3.85} & \textcolor{blue}{\bf 0.0089} & \textcolor{blue}{\bf 2.09} & \textcolor{blue}{\bf 43.60}\\
            \cmark    & \cmark  & \cmark  & \textcolor{red}{\bf 0.989} & \textcolor{red}{\bf 3.85} & \textcolor{red}{\bf 0.0087} & \textcolor{red}{\bf 2.01} & \textcolor{red}{\bf 43.80}\\
            \hline
            \hline
        \end{tabular}
        \label{tab:ablation_multiscale}
    \end{table}
    \begin{figure}[tb]
        \centering
        \includegraphics[width=\linewidth]{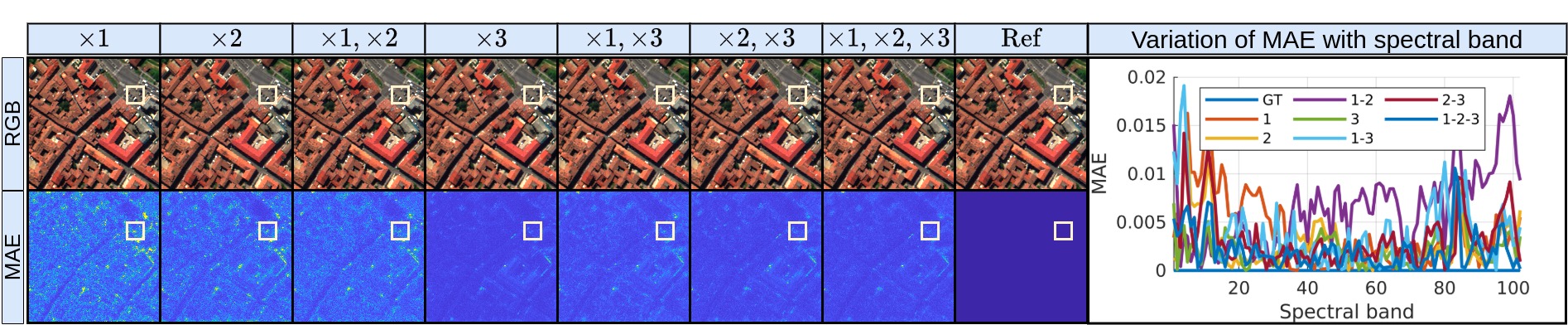}
         \vskip-10pt\caption{The visual results for the ablation study on the effect of HyperTransformer at multiple scales on the Pavia Center dataset.}
        \label{fig:ablation_multiscale}
    \vspace{-5mm}
    \end{figure}
    \vspace{-2mm}
    \paragraph{Multi-scale feature fusion.} As we discussed previously, the conventional pansharpening algorithms usually fuse PAN and LR-HSI features only at a single spatial-scale (i.e., $\times 4$). Instead, our proposed pansharpening network injects textural features from PAN at three different scales, namely $\times 1$, $\times 2$, and $\times 4$ of spatial resolution of LR-HSI. To demonstrate the reason for utilizing HyperTransformers at multiple spatial-scales, we conduct an ablation study, and the results are presented in Table \ref{tab:ablation_multiscale} and Figure \ref{fig:ablation_multiscale}. In Table \ref{tab:ablation_multiscale}, B/L corresponds to the case where no PAN features are injected to the backbone at any spatial-scale (i.e., sharpening without PAN image). As can be seen from Table \ref{tab:ablation_multiscale}, when we inject PAN features to the backbone, the quality of pansharpened HSI improves, and a significant improvement can be noticed when we inject PAN features to the backbone at HR scale (i.e., $\times 4$). Furthermore,  as can be seen from the final row of Table \ref{tab:ablation_multiscale}, the best pansharpening performance is observed when we utilize HyperTransformers across all the three spatial-scales. Concretely, we observe an improvement of $\sim$ 0.3/3.7/11.2/9.0/2.1\% in CC/SAM/RMSE/ERGAS/PSNR  when we utilize HyperTransformers at all scales compared to HyperTransformer only at the $\times 4$ spatial-scale. In addition to the quantitative results, we also present synthesized RGB images, MAE plots, and the variation of MAE with spectral bands for a randomly selected region in Figure \ref{fig:ablation_multiscale}. All the above observations verify that  the multi-scale feature fusion is better than the conventional single-scale feature fusion for HS pansharpening.
    \vspace{-0.5cm}
    \begin{table}[tb]
        \centering
        \caption{Ablation study on different loss functions.\vspace{-3mm}}
        \begin{tabular}{p{0.03\linewidth}p{0.08\linewidth}p{0.08\linewidth}|p{0.06\linewidth}p{0.06\linewidth}p{0.09\linewidth}p{0.10\linewidth}p{0.1\linewidth}}
            \hline
            \hline
            $L_1$  &  $\mathcal{L}_{\text{vgg-per}}$ & $\mathcal{L}_{\text{t-per}}$ & CC & SAM & RMSE  & ERGAS & PSNR\\
            \hline
            \hline
            \cmark    & \xmark           & \xmark           & \bf 0.987 & \bf 4.07 & \bf 0.0091 & \bf 2.12 & \bf 43.00\\
            \cmark    & \cmark  & \xmark           & \textcolor{blue}{\bf 0.988} & \textcolor{blue}{\bf 4.01} & \textcolor{blue}{\bf 0.0090} & \textcolor{blue}{\bf 2.09} & \textcolor{blue}{\bf 43.60}\\
            \cmark    & \cmark  & \cmark  & \textcolor{red}{\bf 0.989} & \textcolor{red}{\bf 3.85} & \textcolor{red}{\bf 0.0087} & \textcolor{red}{\bf 2.01} & \textcolor{red}{\bf 43.80}\\
            \hline
            \hline
        \end{tabular}
        \vspace{-4mm}
        \label{tab:ablation_losses}
    \end{table}
    \paragraph{VGG perceptual loss and Transfer perceptual loss.} Table \ref{tab:ablation_losses} shows how each loss function improves the quality of the  pansharpened HSI. Combining the synthesized perceptual loss $\mathcal{L} _ {\text{vgg-per}}$ with $L_1$ loss improves  CC/SAM/RMSE/ERGAS/PSNR metrics by $\sim $ 0.1/1.5/1.1/1.4/1.4\%, respectively. It is further improved by the transferred perceptual loss  $\mathcal{L}_{\text{t-per}}$ in CC/SAM/RMSE/ERGAS/PSNR by $\sim$ 0.1/4.0/3.3/3.8/0.5 \%, respectively. Note that the significant improvement of PSNR by $\mathcal{L} _ {\text{vgg-per}}$, and improvement of SAM/RMSE/ERGAS metrics by $\mathcal{L}_{\text{t-per}}$.
    
    More experimental results and analysis of the proposed method can be found in the supplementary document.
\vspace{-3mm}   
\section{Limitations and Future Work}
\vspace{-2mm}
From the MAE figures, a relatively high MAE can be observed around UV ($\sim$ bands 1-10) and IR ($\sim$ bands 90-104) regions. This could be due to the lack of UV and IR features in the PAN image. Additional research must be conducted to improve the performance in these regions.
\vspace{-3mm}
\section{Conclusion}
\vspace{-2mm}
In this paper, we proposed a novel textural-spectral feature fusion network for HS pansharpening called HyperTransformer, which transfers HR textural features from PAN image to spectral features from LR-HSI through a multi-head feature soft-attention mechanism. The proposed HyperTransformer consists of two separate feature extractors to extract PAN and LR-HSI features, a multi-head feature soft-attention network to capture the long-range and cross feature-space dependencies between PAN and LR-HSI, and a textural-spectral feature fusion module to fuse HR texture features and spectral features effectively. Furthermore, the proposed HyperTransformer can be utilized in multiple spatial scales to learn more powerful texture representations. Extensive experiments conducted on three widely used HSI datasets demonstrate the superiority of our HyperTransformer over the SOTA approaches on both quantitative and qualitative evaluations. 
\vspace{-2mm}
\section{Acknowledgment}
\vspace{-2mm}
This work was supported by NSF CARRER award 2045489.

\clearpage
{\small
\bibliographystyle{ieee_fullname}
\bibliography{egbib}
}

\end{document}


\title{Supplementary Material for HyperTransformer: A Textural and Spectral Feature Fusion Transformer for Pansharpening}  

\maketitle
\thispagestyle{empty}
\appendix
\begin{figure}[H]
    \centering
    \includegraphics[width=\linewidth]{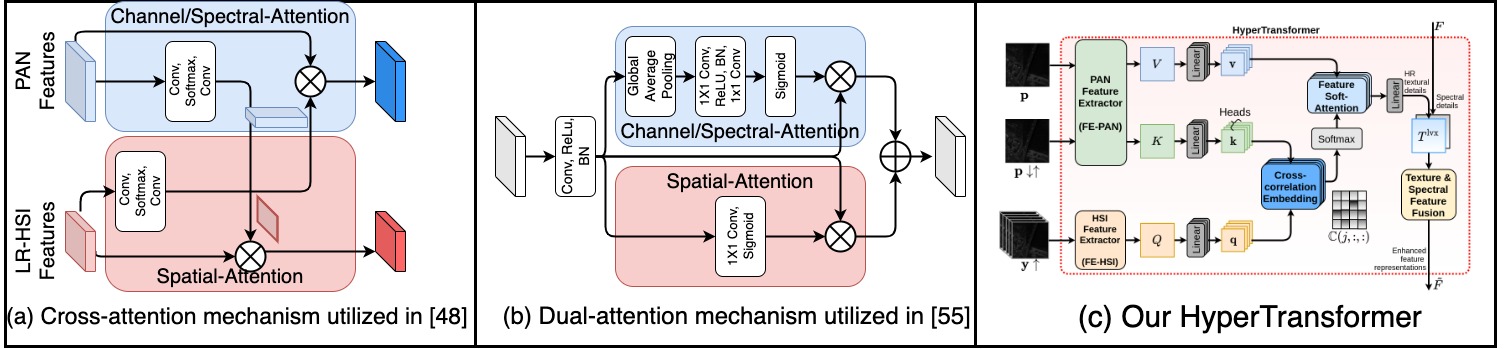}
  \vskip-10pt  \caption{How our proposed \textit{HyperTransformer} differs from the attention mechanisms utilized in previous pansharpening works.}
    \label{fig:attention_comp}
\end{figure}
\paragraph{How our proposed HyperTransformer differs from previous pansharpening methods that utilize attention?}
In previous pansharpening works [48, 55], attention (channel and spatial) mechanisms are used to \textit{re-weight} the PAN and LR-HSI features from ConvNets along the channel and spatial dimensions as shown in Fig. \ref{fig:attention_comp} (a) and (b) without having an explicit consideration of special properties of PAN and LR-HSI features. Different from these previous methods, the proposed HyperTransformer is specifically designed to cater to the pansharpening problem by taking into consideration spatial and spectral properties of PAN and LR-HSI. Instead of simply re-weighting the feature maps, our HyperTransformer first computes the cross-correlation between the feature representations of PAN$\downarrow \uparrow$ and LR-HSI. Then multi-head feature soft-attention (MHFA) is utilized to identify texturally advanced and spectrally similar feature representations from PAN that will be further mixed with spectral features from the backbone network. Hence, the proposed HyperTransformer re-defines queries, keys, and values in standard attention mechanisms as LR-HSI, PAN$\downarrow \uparrow$, and PAN features, respectively that not only deliver better intuitive understanding to the pansharpening problem under the context of attention but also result in better pansharpening performance.  HyperTransformer outperforms many previous classical [1, 2, 3, 16, 18, 23, 25, 33, 40, 41, 52], ConvNet-based [5, 12, 22, 44, 46], and attention-based [55] methods in terms of CC, SAM, RSNR, ERGAS, and PSNR on three datasets.
\begin{figure}[H]
    \centering
    \includegraphics[width=\linewidth]{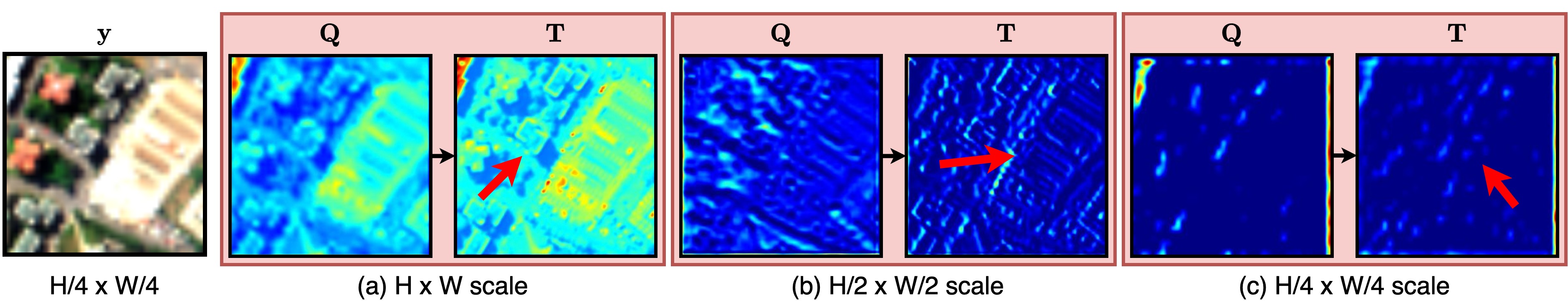}
    \caption{For a given query feature $\bf Q$, we visualize corresponding spectrally similar and texturally advanced feature map $\bf T$ processed from our \textit{HyperTransformer} at multiple spatial scales.}
    \label{fig:features}
\end{figure}
\paragraph{Visualization of \textit{in} and \textit{out} feature maps from our HyperTransformer.}
As shown in Fig. \ref{fig:features}, at each spatial scale,  HyperTransformer adds missing texture details to LR-HSI features (queries - $\bf Q$) while maintaining their spectral characteristics (i.e., the cross-correlation).
\begin{figure}[]
    \centering
    \includegraphics[width=\linewidth]{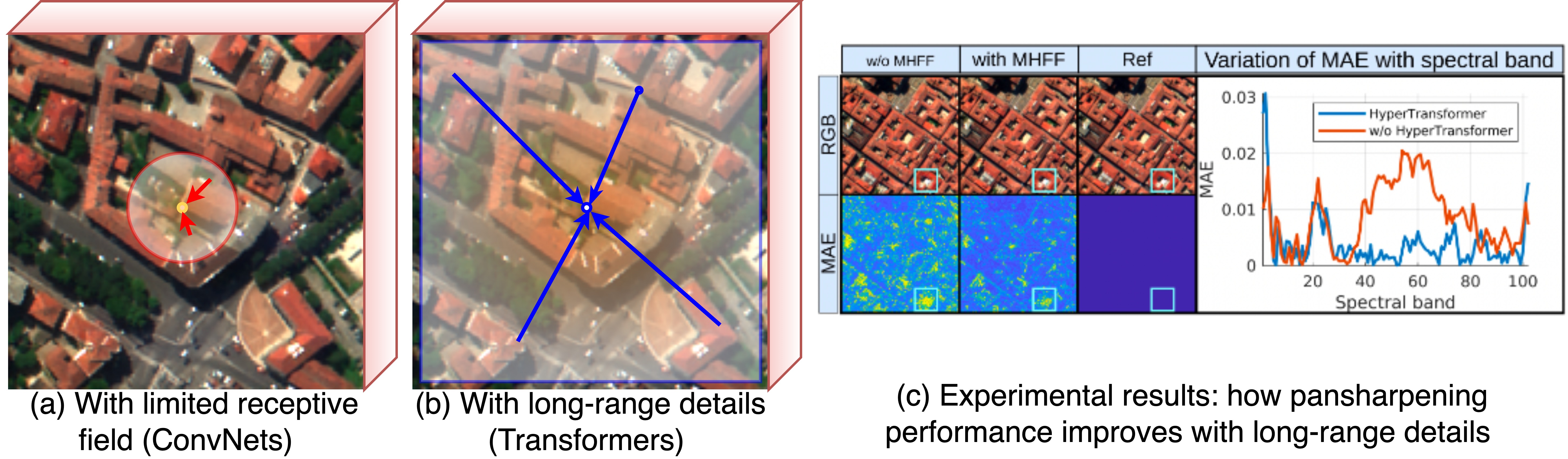}
    \caption{Necessity of long-range details for pansharpening.}
    \label{fig:long_range}
\end{figure}
\paragraph{Why are long-range details necessary for pansharpening?} We explain our intuition of why pansharpening should benefit from long-range details in Fig. \ref{fig:long_range}. As shown in Fig. \ref{fig:long_range}, when the pansharpening network has a larger receptive field (i.e., it can capture long-range details), it can enhance the texture and spectral details of a given pixel not only by looking at adjacent pixels but also from the pixels far away. As shown in \ref{fig:long_range} - (c) (in paper Fig. 4), we can see a significant reduction in MAE across the spectral bands when we add our HyperTransformer to the main pansharpening network, which empirically shows that  pansharpening indeed benefits from long-rage details. Furthermore, it has been shown in the literature that not only segmentation, detection, and classification tasks benefit from long-rage details but also restoration, fusion, and super-resolution \cite{transformer_survey}.  Pansharpening consists of both super-resolution and fusion tasks and as a result it should also benefit from long-range details.

{\small
\bibliographystyle{ieee_fullname}
\bibliography{egbib}
}
\clearpage